\documentclass{article}

\usepackage[margin=1.2in]{geometry}
\usepackage{graphicx}
\usepackage{float}
\usepackage{amsmath}
\usepackage{amssymb}
\usepackage{natbib}

\usepackage{graphicx}
\usepackage{float}
\usepackage{amsmath}
\usepackage{amssymb}
\usepackage{natbib}
\usepackage{wrapfig}

\makeatletter
\newsavebox{\@brx}
\newcommand{\llangle}[1][]{\savebox{\@brx}{\(\m@th{#1\langle}\)}%
	\mathopen{\copy\@brx\kern-0.5\wd\@brx\usebox{\@brx}}}
\newcommand{\rrangle}[1][]{\savebox{\@brx}{\(\m@th{#1\rangle}\)}%
	\mathclose{\copy\@brx\kern-0.5\wd\@brx\usebox{\@brx}}}
\makeatother

\newcommand{\bra}[1]{\left\langle #1 \right|}
\newcommand{\ket}[1]{\left|#1\right\rangle}

\title{Quantum Bayesian Computation}

\author{	
	\makebox[.4\linewidth]{Nick Polson\footnote{Nick Polson is Professor at Chicago Booth: ngp@chicagobooth.edu  Vadim Sokolov is an Assistant Professor at Operations Reaearch at George Maon U: vsokolov@gmu.edu Jianeng Xu is at Chicago Booth: jianeng.xu@chicagobooth.edu }}\\
	\textit{  Booth School of Business}\\
	\textit{  University of Chicago}\\
	\and 
	\makebox[.4\linewidth]{Vadim Sokolov}\\
	\textit{ Department of Systems Engineering }\\
	\textit{  and Operations Research}\\
	\textit{ George Mason University}\\
	\and
	\makebox[.4\linewidth]{Jianeng Xu}\\
	\textit{ Booth School of Business}\\
	\textit{  University of Chicago}\\
}

\graphicspath{{fig/}}
\begin{document}
\maketitle

\begin{abstract}
\noindent Quantum Bayesian Computation (QBC) is an emerging field that levers the computational gains available from quantum computers 
to provide an exponential speed-up in Bayesian computation.  Our paper adds to the literature in two ways. First, we show how von Neumann quantum measurement can be used to 
simulate  machine learning algorithms such as Markov chain Monte Carlo (MCMC) and Deep Learning (DL) that are fundamental to Bayesian learning. 
Second, we describe data encoding methods needed to implement quantum machine learning  including  the counterparts to traditional feature extraction and kernel embeddings methods.
Our goal  then is to show how to  apply quantum algorithms directly to statistical  machine learning problems.  On the theoretical side, we provide  quantum 
versions of high dimensional regression, Gaussian processes (Q-GP)  and stochastic gradient descent (Q-SGD).
On the empirical side, we  apply a Quantum FFT model to Chicago housing data. Finally, we conclude with directions for future research.
\end{abstract}

\vspace{0.15in}
    
Key Words:  Quantum Bayes, Quantum Data Encoding, Quantum MCMC, Quantum Deep Learning,  Quantum Feature Extraction, Quantum Embedding, Quantum Learning, Quantum Sensing, Quantum Entanglement, Quantum Superposition, Quantum Encryption, Quantum FFT, Quantum Regression, Quantum Gaussian Processes, 
Quantum Stochastic Gradient Descent (Q-SGD), von Neumann Quantum Measurement.

\section{Introduction}
Quantum Bayesian Computation (QBC)  holds the promise of gains in computational speed that are available from quantum computers.
Quantum computers are devices that harness quantum mechanics to perform computations in ways that classical computers cannot. 
Quantum algorithms  promise is an exponential speed-up for Bayesian machine learning. 

Quantum state measurement is based on an idea of \cite{von2018mathematical} where the state of a physical system is used
to measure quantities of interest, von Neumann's  principle of quantum measurement is central to quantum measurement and simulation.
\cite{feynman1986quantum}, in a seminal paper,  discusses the implications of quantum physics for mechanical computational methods.  From a statistical viewpoint,
\cite{smith1984present} and \cite{Breiman2001two} provide a framework for Bayesian computation and black box neural networks for data science in the 21st century. 
Whilst the quantum mechanics underlying the construction of a quantum computer is still in its infancy, there are many algorithms that are available for speed-up of existing methods.

Simulating a quantum system has counterparts in Bayesian learning such as MCMC simulation and Deep Learning.
For example, we provide a version of quantum deep learning (Q-DL) achieves this via a superposition of semi-affine functions whereas quantum computers use the Bloch sphere and a variety of quantum embeddings. Recent reviews of quantum computing and its promise for big data analytics are \cite{Wang2022When,hidary2021quantum} and \cite{schuld2019Quantum,schuld2021Supervised}.

Quantum embedding is an important part of the data encoding process and the type of functions that are easy to evaluate with quantum simulation.
In particular, we show that  quantum embeddings can be viewed as traditional  kernel methods with a specific data encoding, 
First, we create a density matrix of states of size $2^b \times 2^b $ from a $b$ qubit computer. This will form the equivalent of the kernel space that is commonplace in learning methods.
Machine learning methods essentially use linear algebra algorithms on this augmented feature space and quantum computers are directly applicable to such storage and calculations as they typically provide an exponential speed up in such algorithms, e.g. principal components, matrix inversion, Fourier transforms, inverse problem to a name a few.

The fundamental problem of machine learning is to reconstruct input-output representations. To do this, we first represent them using quantum interaction terms. 
Then we use a quantum computer to simulate the solution to Schro\"dinger's equation to provide the von Neumann measurement of the quantity of interest.
Turing showed that you can represent an $n$-dim function (boolean) $ \{0,1\}^n \rightarrow \{0,1\}$. In $ \lambda $-calculus and evaluate it as a composite function. 
Kolmogorov-Arnold  theory discusses the representation of multivariate functions.
Essentially, qubit activation functions allow you to represent Boolean functions and compute then quickly. 
Much of the gains then  from quantum computing will be in regard to computability and representation of multivariate functions.  

The speed of function evaluation is of paramount importance  to artificial intelligence and lies at the heart of the promise of quantum computers.  Quantum supremacy has already been demonstrated in a number of experiments.  From a theoretical viewpoint, \cite{church36unsolvable} defines the so-called $ \lambda $-calculus where $ \lambda $ denotes a function.
\cite{turing1937Computability} famously wrote about computability of functions (a.k.a predictive rules).  The promise of quantum computation and physical measurement by simulation relies on the simple observation that it only takes a few dozen qubits to be able to calculate a large superposition of functions. Compositions of functions (as opposed to additive structures) such as deep learners are also
naturally computable. Computer languages such as Haskell are directly designed to perform such calculations. \cite{polson2020deep} provide a discussion of symbolic differentiation and manipulation for deep learning. we show that gradients are also straightforward to  compute using quantum  von Neumann measurement. 

The rest of our paper is outlined as follows.  Section 1.1 provides a review of  quantum probability ( \cite{wang2011quantum}) and simulation.   Section 2 provides our Quantum Bayesian Computation (QBC) framework. We show how  von Neumann's principle of
quantum measurement  can be used to implement MCMC and DL functions after suitable data encoding transformations. Section 3 described quantum data encoding methods.
Section 4 provides quantum algorithms specifically designed to tackle problems in machine learning such as high dimensional regularized regression, Gaussian process regression
and quantum stochastic gradient descent (Q-SGD).
Appendix A contains commonly used  quantum algorithms. Finally, we conclude with directions for future research.
We begin with a basic review of classical and quantum probability to contrast the two approaches.

\subsection{Quantum Bayes Probability}

This subsection contrasts classical and quantum probability and describes quantum entanglement and superposition.

\vspace{0.1in}

\noindent{\bf Classical Probability.} In a simple coin tossing experiment, $y$ takes one of two values $ \{ 0 ,1 \} $ with probability $ p_1 , p_0 $ where $ p_1 = 1 - p_0 $. Here 
$ p_0, p_1 > 0 $ and $ \lVert p \rVert_1 =1 $. 

\vspace{0.1in}

\noindent{\bf Quantum Probability.} Quantum probability generalises this to the complex plane with the $L^2$-norm for its complex coefficients. One can view this as a data transformation to the Bloch sphere which is shown in Figure 1.

\begin{figure}[H]
\centering
\includegraphics[width=0.3\linewidth]{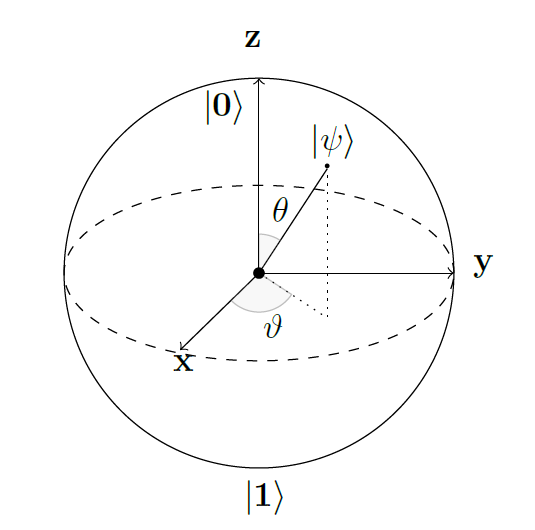}
\caption{Bloch sphere}
\label{fig:Bloch}
\end{figure}

\paragraph{\bf Qubit} Like a bit is a basic unit of information in traditional computing, The qubit is a basic quantum unit and it encodes a two-state quantum system, such as spin of the electron (up/down). We can measure a qubit (the simplest quantum system) and obtain either $ 0$ with probability $  \lVert a_0 \rVert^2 $ or $ 1$ with 
probability $  \lVert a_1 \rVert^2 $.
Using Dirac's notation, besides states $ | 0 \rangle $ and $ | 1 \rangle $, a qubit can also take states as their super-positions, which are linear combinations of the form 
\[
| \psi \rangle = a_0 | 0 \rangle + a_1  | 1 \rangle
\]
where $ a_0 , a_1 \in \mathbb{C} $ are called amplitudes such that $  \lVert a_0 \rVert^2 +   \lVert a_1 \rVert^2 = 1 $.
The notation $ | + \rangle = \frac{1}{2} | 0 \rangle +   \frac{1}{2} | 1 \rangle $ is used for the halfway state generated by superposition.

Qubits can evaluate boolean functions quickly particularly deep learning representations based on superposition of affine functions. 
In general, a system of $b$ qubits has $2^b$ computational basis states of the form $ | x_1 x_2 \ldots x_b \rangle $ where $x_j =0,1$ and $ 1 \leq j \leq b $ that generate a $2^b$-dim complex vector space and a superposition state that is based on $2^b$-amplitudes. Hence, we only need a system of a few dozen 'qubits' to provide a significant increase in computational speed as $ 2^{52} \approx 10^{16} $. A system with $b=500$ qubits would have the ability to model more than the number of atoms in the universe.

\paragraph{Quantum Entanglement.} Quantum entanglement is the physical phenomenon that occurs when a group of particles are generated, interact, or share spatial proximity in a way such that the quantum state of each particle of the group cannot be described independently of the state of the others, including when the particles are separated by a large distance. The topic of quantum entanglement is at the heart of the disparity between classical and quantum physics: entanglement is a primary feature of quantum mechanics lacking in classical mechanics.

\paragraph{Quantum Superposition} An intrinsic difference exists between  bit representation in classical and quantum systems. All classical algorithms prevent the simultaneous occurrence of their states. The quantum system allows for simultaneous occurrence of their states. The computational basis takes the form $| x_1,x_2,\ldots,x_b\rangle$, where $x_j={0,1},~j=1,\ldots,b$. Any quantum superposition state is a liner combination of $2^b$ possible base states whose complex coefficients are called amplitudes. A $b$-qubit state $| \psi \rangle$ may take u superposition state of the form
\[
    | \psi \rangle = \sum_{x_1,x_2,\ldots,x_b}\alpha_{x_1,x_2,\ldots,x_b}| x_1,x_2,\ldots,x_b\rangle, \quad \mathrm{s.t. } \sum_{x_1,x_2,\ldots,x_b}|\alpha_{x_1,x_2,\ldots,x_b}|^2 = 1.
\]

\section{Quantum Bayesian Computation (QBC)}

\noindent{\bf Bayesian Machine Learning.}  Supervised statistical learning given data as a sequence of input-output pairs. The goal is to determine a predictive rule for new input cases or to describe with statistical uncertainty is described by  predictive distribution $p(y\mid x)$ of future observations. 

How does one achieve such a goal given the empirical data of $N$ input-output pairs $(y_i,x_i)_{i=1}^N$. We need to formulate an input-output map, $y = f(x)$, the question is how to construct $f$?

The machine learning problem then is to find a good predictor rule from a training dataset if input-output pairs $(y_i,x_i)_{i=1}^N$ of observed data. 
The goal is to predict at a new high dimensional $ x_i =  ( x_{1i} , \ldots , x_{pi} ) $.
To achieve good generalisability we need to be able to perform nonlinear dimension reduction and to find a suitable set of features/factors. 
Hence, machine learning algorithms rely on interpolation. Gaussian processes and deep learners provide two classes of interpolators that have been shown to do well in many fields of application. We will provide quantum speed-ups of gaussian processes and a framework for quantum deep learning. An area of future research, is whether quantum representations provide other useful classes of interpolators.

Given a set of $(output, input)$ pairs $ (y_i,x_i)_{i=1}^N $, the goal is simply to find a mapping (a.k.a. data transformation), denoted by $ \hat{f} $, where
$$
y_i = f (x_i ) 
$$
and then to find a generalised prediction rule, that holds $ \forall x \in \mathcal{X} $, denoted by 
$$
\hat{y} ( x) = \hat{f}( x ) 
$$
Here $ \hat{f} $ is estimated from the "training" dataset.

Bayesian predictive calculations  start with a probabilistic model and map (a.k.a. data generating process), denoted by $ p( y | x ) $. 
Under predictive mean squared error (MSE), the optimal rule is then given by the predictive mean
$$
\hat y(x) = E(y\mid x) = \sum_{y \in Y} y \; p( y| x ) 
$$
The use of latent (hidden) variables, denoted by  $z$, is also an important feature of Bayesian thinking.
They can be viewed as "extending the art of the conversation", see \cite{Lindley1990}.  In quantum computing they can be viewed as states of a physical system.
With latent variables, $z$, we can write, the optimal predictive rule as  
$$
\hat y(x) = E \left [ E(y\mid x,z) \right ] .
$$
This is simply the law of total probability. The key insight is that it is easier to assess the inner expectations, the caveat being that we need to evaluate large sums which are tailored made for quantum computers.
This includes the case of classification, where we need to evaluate classification probabilities $  p(y=1\mid x) $ via  expectations of an indicator functions.

The predictive distribution is given by
$$
p( y|x) =\frac{  p( x|y) p( y) }{p(x)}
$$
The introduction of latent states, $z$, allows the posterior  to be broken into simpler parts
$$
p( y|x) = \sum_{z \in Z} p(y|z,x) p( z|x) 
$$
Again we require enumeration over the latent states $z$.  These expectations can be calculated via von Neumann's quantum  measurement simulation as a state of a system.  
This is similar to the view of a particle filter where the researcher simulates a stochastic process whose marginals over time as realisations of the sequence of posterior distributions
that one wishes to calculate.

\subsection{von Neumann Quantum Measurement}

von Neumann's key idea was  to 'measure' the ensemble sum via the simulation of the physical system. A quantum state is completely described by density operators. A 
vector $ | \psi \rangle \in \mathbb{H} $ corresponds to a density operator $ \rho = | \psi \rangle \langle \psi | $ the projection onto $ | \psi \rangle $.
An ensemble state $ \rho_k $ with probability $p_k$ of being in vector $ | \psi_k \rangle $ has density operator
\[
 \rho_k = \sum_{k=1}^K p_k | \psi_k \rangle \langle \psi_k | 
\]
Suppose that $ \rho $ is generated by a $b$-qubit state, creating a $ 2^b \times 2^b $-Hermitian density matrix. 
This allows us to construct a general paradigm for quantum measurement in machine learning.

\vspace{0.1in}

\noindent{\bf Quantum Simulation} Here the initial complex state is subject to a unitary transformation
\[
| \psi^{(t)} \rangle = U^t | \psi^{(0)} \rangle
\]
where $ U = e^{- i H }$ is a unitary operator with $ U U^\dagger = 1 $ and $H$ is the Hamiltonian of the system.

In continuous-time we can argue as follows. The heat equation simulation is to solve Schr{\"o}dinger's equation 
\[
i \frac{ \partial | \psi^{(t)} \rangle}{ \partial t} = H | \psi^{(0)} \rangle
\]
starting from initial state $ | \psi( 0) \rangle $ with solution
\[
| \psi( t) \rangle = e^{- i H t} | \psi( 0) \rangle
\]
The numerical evaluation of $ e^{-i Ht} $ is needed and this requires discretization using Trotter's formula (Trotter, 1959).

\vspace{0.15in}
\noindent{\bf Quantum MCMC Simulation} Here the initial probabilistic state $ p^{(0)} $ is subject to a stochastic transition matrix, $P$, that is time reversible and the evolution of the system is given by
\[
p^{(t)} = P^t p^{(0)}
\]
In continuous-time $ P$ can be generated by the infinitesimal generator (which takes the place of the Hamiltonian in the quantum system)

\vspace{0.15in}
\noindent{\bf Quantum Deep Learning}
Here the initial state $ x^{(0)} $ is subject to an iterate map (with $t$-layers)
\[
x^{(t)} = F^t x^{(0)}
\]
where the map $ F$ is. a composition of ridge functions (see \cite{polson2021deep}).

\vspace{0.1in}
With a quantum state prepared in a physical state, $ \rho $, the observable $Y$ with values in $ \{ y_1 , \ldots , y_M \} $ has a probability distribution
\[
p_\rho ( Y = y ) = tr ( \rho Q_y ) = \sum_{j=1}^M y_j p_\rho ( Y = y_j )
\]
The equivalence between everything can be thought of as an iterative map (a.k.s. stochastic simulation)

Hence, from the von Neumann measurement viewpoint, all there methods can be though of as symbolic manipulation of linear operators leading to the physical evaluation 
of the quantity of interest.
The main advantage of quantum systems is the 
exponential speed up as you can directly calculate the sums need for Bayesian learning, Expectations and Predictive rules which we discuss in Section 3. 
Deep learners have the advantage of representing multivariate functions.

\subsection{Quantum Feature Extraction}

From a data analytic and Bayesian learning perspective, one can think of the mapping $ \rho : x \rightarrow \rho(x)$ as a fixed feature map providing a data encoding from $ \mathcal{X} $ to density matrixes. This quantum embedding is the equivalent of feature selection in machine learning (see \cite{hoadley2001statistical},\cite{bhadra2021merging}, \cite{polson2021deep}) and is an important part of which class of functions that can be represented by quantum neural networks, The corresponding quantum kernel is
$$
k( x , x^\prime ) = tr ( \rho (x) \rho( x^\prime ) )
$$
with the space of functions, for weight operator $W$ being of the form 
$$
f_W ( x )=  tr( \rho (x) W )
$$
For example, Kernel ridge regression $ y =  f_W (x) + \epsilon $ will be straightforward to calculate using quantum simulation and von Neumann measurement.

Specifically, if $ | \psi (t) \rangle $ is the state of the quantum system, then  $ | \psi (t_1) \rangle $  and  $ | \psi (t_2) \rangle $ are related by
$$ 
| \psi (t_2) \rangle = U(t_1 , t_2 )  | \psi (t_1) \rangle  \; \; {\rm with} \; \; U(t_1 , t_2 ) = e^{i H ( t_2 - t_1 )}
$$ 
where $U$ is a unitary operator and $H$ is a Hamiltonian. In practice this operator has to be discretized in simulation via Trotter's \citep{trotter1959} formula. The Trotter's formula approximates $e^{-iH\delta}$ by $U_{\delta}$, which requires only the evaluation of each $e^{-iH_l\delta}$, where
\[
U_{\delta} = [e^{-iH_1\delta/2},\ldots,e^{-iH_L\delta/2}][e^{-iH_1\delta/2},\ldots,e^{-iH_L\delta/2}],
\]
see also \cite{wang2011quantum}.

\section{Quantum Data Encoding}
As data grows in scale and complexity and algorithms such as deep learning become more elaborate, computational techniques become ever more important.

\begin{figure}[H]
\centering
\includegraphics[width=\linewidth]{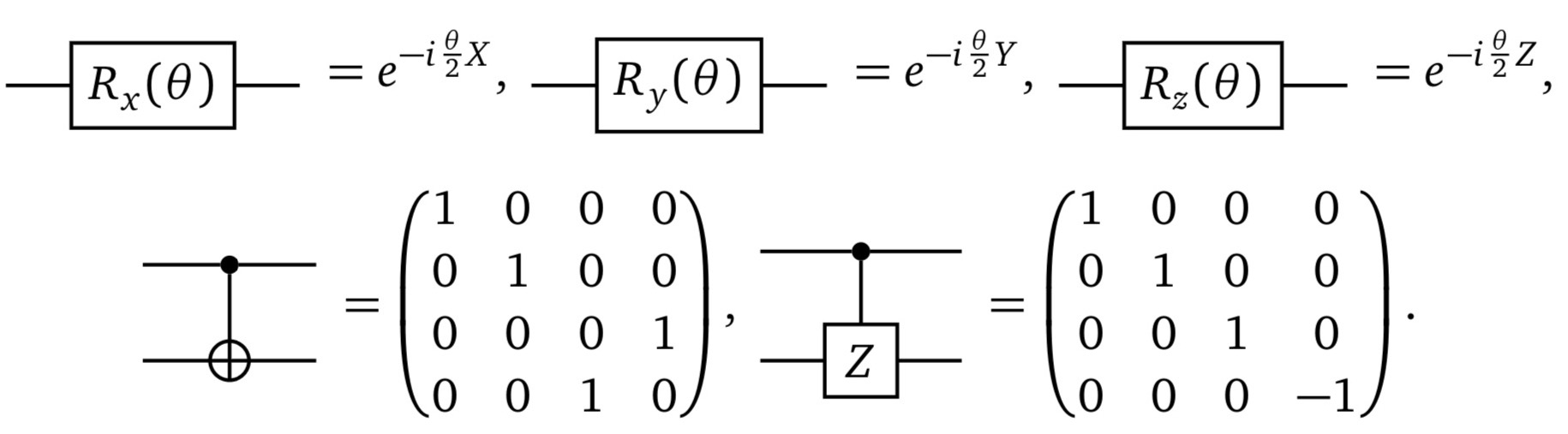}
\caption{Quantum Gates}
\label{fig:gates}
\end{figure}

\paragraph{Quantum Gates and Circuits.} 
There are many gates and circuits that can be used to transform quantum probabilities on the Bloch sphere.
For example, one of the simplest gate is a quantum NOT gate maps $|0\rangle$ to $|1\rangle$ and $|1\rangle$ to $|0\rangle$ and transforms $|\psi\rangle = \alpha_0|0\rangle + \alpha_1 |1\rangle$ into $|\psi\rangle = \alpha_0|1\rangle + \alpha_1 |0\rangle$. The quantum NOT gate can be represented by the Pauli gate matrix
\[
    \begin{pmatrix}
        1 & 0 \\
        0 & 1
    \end{pmatrix},
\]
which has the property of being a unitary operator in $\mathbb{C}^2$. Figure \ref{fig:gates} shows other quantum gates. 

Another useful data Encoding gate is the rotational gate. Here the 
encoding gate $U_{\rho}(x)$ transforms data samples $x=(x_1,\ldots,x_n)\in \mathbb{R}^n$ into a quantum state 
\[
|x\rangle = U(x)|0\rangle = R_x(x_1)\circledcirc R_x(x_2),\ldots,R_x(x_n)|0\rangle^{O\times n},
\]
where 
\[
R_x(x) = e^{-i\phi \sigma_x/2} = 
\begin{pmatrix}
    \cos (\phi/2) & -i\sin (\phi/2) \\
    -i\sin (\phi/2) & \cos (\phi/2)
\end{pmatrix}
\]
We will discuss the application of these later in the our discussion of Quantum Neural Networks. Now we describe quantum entanglement.

\subsection{Quantum Kernels}

Similar to traditional  statistical learning methods, quantum learning can be viewed as a  data map  into a high-dimensional feature space, they can be trained and selected in a low-dimensional subspace. With the density matrices $\rho(x)$ given by some embedding as feature vectors, quantum machine learning models are linear in the feature space. And borrowing ideas from classical machine learning kernel theory, similar results can be derived from the quantum kernel $k(x, x') = tr(\rho(x')\rho(x))$. For example, the optimal models that minimize the cost is linear expansion in terms of quantum kernel functions:
$$
f(x) = \sum_{m=1}^{M} \alpha_m tr(\rho(x^m)\rho(x))
$$
where $x^m, m=1,..., M$ are training data. Hence searching for the optimal model based on a given dataset is essentially an $M$-dimensional optimisation problem. Furthermore, by choosing a convex loss function, this help avoid the barren plateaus in variational training and guarantees the optimality.

Quantum kernel, the inner product of two feature vectors $\rho(x')$ and $\rho(x)$, depends on the underlying data-encoding $\phi(x)$: 
$$
k(x, x') = tr(\rho(x')\rho(x)) = |\langle \phi(x') | \phi(x) \rangle|^2.
$$
The simple amplitude encoding is defined as 
$$
\phi(x) = |x \rangle \langle x| = \sum_{i,j=1}^{N} x_i x_j^* |i \rangle \langle j|
$$
where $x$ is normalized such that $\Vert x \Vert^2 = \sum_i x_i^2 = 1$ and $i, j$ are computational basis. The induced kernel is $k(x, x') = |\langle x' | x \rangle|^2 = |x^\dagger x'|^2$.

Other interesting and common strategies including: basic encoding
$$
\phi(x) = |\langle i_x' | i_x \rangle|^2 
$$
which gives the Kronecker kernel $k(x, x') = \delta_{x, x'}$ where the input binary string $x$ is represented by the integer $i_x = \sum_{k=0}^{n-1} 2^k x_k$, and coherent state encoding which gives a Gaussian kernel. There is resemblance between quantum kernels and classical machine learning kernels, and many of them have Fourier representation.

A quantum model is then defined as the expectation of the measurement $\mathcal{M}$ under the density matrix $\rho(x)$, 
$$
f(x) = tr(\rho(x) \mathcal{M})
$$
and the measurement $\mathcal{M}$ can always be expressed as a linear combination of $\rho(x^k)$ where $x^k$ are from the data domain. More specifically, there exist $\{x^k, k=1,2,...\}$, such that
$$
tr(\rho(x) \mathcal{M}) = tr(\rho(x) \mathcal{M}_{exp})
$$
for all $x$, where $\mathcal{M}_{exp} := \sum_k \gamma_k \rho(x^k) $. More importantly, this indicates that any quantum model can be represented as
$$
f(x) = tr\left(\rho(x) (\sum_k \gamma_k \rho(x^k) )\right)= \sum_k \gamma_k k(x^k, x)
$$

\subsection{Optimizing Quantum Models}
The representer theorem states that the optimal quantum models which minimize the regularized empirical risk,
$$
\frac{1}{M} \sum_{m=1}^M L(y^m, f(x^m)) + \gamma \Vert f \Vert_F^2
$$
admits a representation of the form
$$
f(x) = \sum_{m=1}^M \alpha_m k(x^m, x)
$$
Here $x^m$ are training samples and $F$ is the reproducing kernel Hilbert space (RKHS) corresponding the the encoding. Hence, the optimization over the regularized empirical risk is the same as finding the optimal measurement $\mathcal{M}$. The vectorized $\mathcal{M}_{opt}$ is the familiar form
$$
|\mathcal{M}_{opt} \rrangle = \sum_m y^m \left(\sum_{m'}  |\rho(x^{m'}) \rrangle \llangle \rho(x^{m'})  | \right)^{-1}   |\rho(x^{m})  \rrangle,
$$ 
This is analogous to the ordinary least square result for regression coefficients 
as quantum models are linear in feature space. 

\subsection{Quantum High Dimensional Regression}
In a high dimensional regression setting, we need to calculate $\hat \beta = X^{\dagger}y$
 where $X^{\dagger}$ is the Moore-Penrose pseudo-inverse of $X$. The singular decomposition can be computed as $X^{\dagger} = V\Sigma^{-1}U^{\dagger}$  and hence represented as a quantum state, see \cite{schuld2016prediction}. Then we need to calculate
\[
\bar \beta = \sum_{k=1}^K \sigma_k^{-1}v_ku_k^Ty
\]
The optimal prediction rule under MSE, is given by 
\[
	\hat y (x) = \sum_{k=1}^K \sigma_k^{-1}x^Tv_ku_k^Ty
\]
We need to calculate a set of inner products with is easily computable using a quantum computer which are naturally designed to calculate inner products.
See \cite{harrow2009quantum} for further discussion.

\subsection{Quantum Gaussian Process Q-GP}
A quantum algorithm for quantum algorithm for Gaussian process regression was introduced in \cite{zhao2019bayesian}. The GP predictor is
\[
\hat f_* = k^T_*(K+\sigma^2 I_n)^{-1}y,
\]
depending on whether the goal is to compute the mean or variance, we choose $|b\rangle = |y\rangle$ or $|b\rangle = |k_*\rangle$, where $k_* = k(x_*,x_*)$ is the covariance of the target point with itself, with the corresponding formula for the predictive variance. The quantum GP algorithm simulates $K + \sigma^2I_n$ as a Hamiltonian acting on an input state $|b\rangle$. Then performs quantum phase estimation to extract estimates of the eigenvalues of $K_\sigma^2I_n$  and stores them as a weighted superposition and then creates the desired prediction. \cite{schuld2021Supervised} show how to provide estimates of 
\[
k_*^T  \left(K + \sigma^2I_n\right)^{-1} y \; \; {\rm and} \; \; 
	k_*^T\left(K + \sigma^2I_n\right)^{-1}k_*,
\]
written as two inner products.

\subsection{Quantum Stochastic Gradient Descent (Q-SGD)}
Gradients are much easier to calculate in quantum systems.  This is due to the following property of unitary transformations.
To implement the gradient descent, we  approximate the derivative of the loss function by taking the symmetric difference,
\begin{equation}
 				\frac{df}{dx}(x) = \big(f(x+\epsilon)-f(x-\epsilon)\big)/(2\epsilon) + O(\epsilon^2).
\label{farhar20b}
\end{equation}
However the gradient can be analytically calculated when we use the following unitary operators
\begin{equation}
\exp\, (i\, \theta\, \Sigma)
\label{farhar7}
\end{equation}
where $\Sigma$ is a generalized Pauli acting on a few qubits, that is, $\Sigma$ is a tensor product of operators from the set $\left\{\sigma_x, \sigma_y, \sigma_z\right\}$ acting on a few qubits.  The derivative with respect to $\theta$ gives an operator whose norm is bounded by 1. Therefore the gradient of the loss function with respect to $\vec{\theta}$ is bounded by $L$, the number of parameters.

For this function the gradient is the generalized Pauli operator $\Sigma_k$
\begin{equation}
  \frac{d \text{loss}(\vec{\theta}, z)}{d \theta_k} = 2 \operatorname{Im}\big( \bra{z , 1} U_{1}^\dagger... U_{L}^\dagger Y_{n+1} U_L ...U_{k+1} \Sigma_k U_k ... U_{1} \ket{z, 1}\big)
\label{farhar20c}
\end{equation}
Note that $Y_{n+1}$ and $\Sigma_k$ are both unitary operators.  Define the unitary operator 
\begin{equation}
   \mathcal{U}(\vec{\theta})=  U_{1}^\dagger... U_{L}^\dagger Y_{n+1} U_L ...U_{k+1} \Sigma_k U_k ... U_{1}  
   \label{farhar20d}
\end{equation}
so we reexpress \eqref{farhar5} as 
\begin{equation}
   \frac{d \text{loss}(\vec{\theta}, z)}{d \theta_k} = 2 \operatorname{Im}\big( \bra{z , 1} \mathcal{U} \ket{z, 1}\big).
   \label{farhar20e}
\end{equation}
$\mathcal{U}(\vec{\theta})$ can be viewed as a quantum circuit composed of $2L+2$ unitaries each of which depends on only a few qubits. We can use our quantum device to let $\mathcal{U}(\vec{\theta})$ act on $\ket{z, 1}$. Using an auxiliary qubit we can measure the right hand side of \eqref{farhar20e}.  To see how this is done start with
\begin{equation}
				\ket{z, 1}  \frac{1}{\sqrt{2}}  \big( \ket{0} + \ket{1} \big)
\end{equation}
and act with $i \mathcal{U}(\vec{\theta})$ conditioned on the auxiliary qubit being 1.  This produces

\begin{equation}
				\frac{1}{\sqrt{2}} \big( \ket{z, 1} \ket{0} + i \mathcal{U}(\vec{\theta}) \ket{z , 1}  \ket{1}\big)
\end{equation}
Performing a Hadamard on the auxiliary qubit gives
\begin{equation}
				\frac{1}{2}\big( \ket{z , 1}  + i \mathcal{U}(\vec{\theta}) \ket{z , 1}  \ket{0} \big)  + \frac{1}{2}\big( \ket{z , 1}  - i \mathcal{U}(\vec{\theta}) \ket{z , 1} \ket{1}  \big) .
\end{equation}
Now measure the auxiliary qubit.  The probability to get $0$ is
\begin{equation}
					\frac{1}{2} - \frac{1}{2} \operatorname{Im}\big( \bra{z, 1} \mathcal{U}(\vec{\theta}) \ket{z, 1}  \big)
\end{equation}
Hence by making repeated measurements we can get a good estimate of the imaginary part which turns into an estimate of the $k$\textsc{\char13}th component of the gradient.  This method avoids the numerical accuracy issue that comes with approximating the gradient as outlined in the previous paragraph.  The cost is that we need to add an auxiliary qubit and run a circuit whose depth is $2L+2$.

\section{Application: Quantum TensorFlow}
We demonstrate an application of a quantum neural network to the problem of image classification. It is assumed that input is an image that is represented as a sequence of $n$ elements $z = (z_1,\ldots,z_n)$, with each $z_i$ being +1 or -1. The output $l(z)$ is also binary and takes values +1 and -1. The basic building block of a quantum neural network is a unitary operator
\[
U_a(\theta).
\]
We assume that this operator acts on a subset of the qbits and is defined by parameter $\theta$. Then the neural network is a composition of those unitary operators.
\[
U(\theta) = U_L(\theta_L)U_{L-1}(\theta_{L-1})\ldots U_1(\theta_1).
\]
Assume our output (readout) bit is 1, then the corresponding input $z$, we construct a computational basis
\[
|z,1\rangle = |z_1,\ldots,z_n,1\rangle.
\]
Applying our neural network to the input $z$ gives the state
\[
U(\theta)|z,1\rangle.
\]
Then on the readout qubit, we measure a Pauli operator $\sigma$ which give us +1 or -1. The goal is that this outcome matches the true label of the data (e.g. image). Given that the outcome is uncertain, we can use multiple copies of the outputs and then average them out.

\vspace{0.1in}

\subsection{Housing Data Application}
For real estate market stakeholders, selling or purchasing a home is an uncertain process. Sellers face uncertainty about how to price their home, how quickly it might sell, and how many offers they might expect from interested buyers in the market. We use the showing events for homes listed for sale in Chicago for a short term housing  demand. The showing data employed spans January 1, 2011 to December 31, 2013 and comprises all homes, condominiums, and apartments listed for sale in Chicago that ShowingTime's scheduling system recorded. The resulting dataset contains 6 million property actions including showings, inspections, open houses, or any other appointment that ShowingTime's scheduling service recorded. However, the records included in the analysis were limited to home showings, which comprised 4.6 million records. The data is further restricted to exclude rental properties, retail, and properties selling bundled units together, which left 3.9 million residential properties.

\begin{figure}[H]
	\centering
	\includegraphics[width=0.5\textwidth]{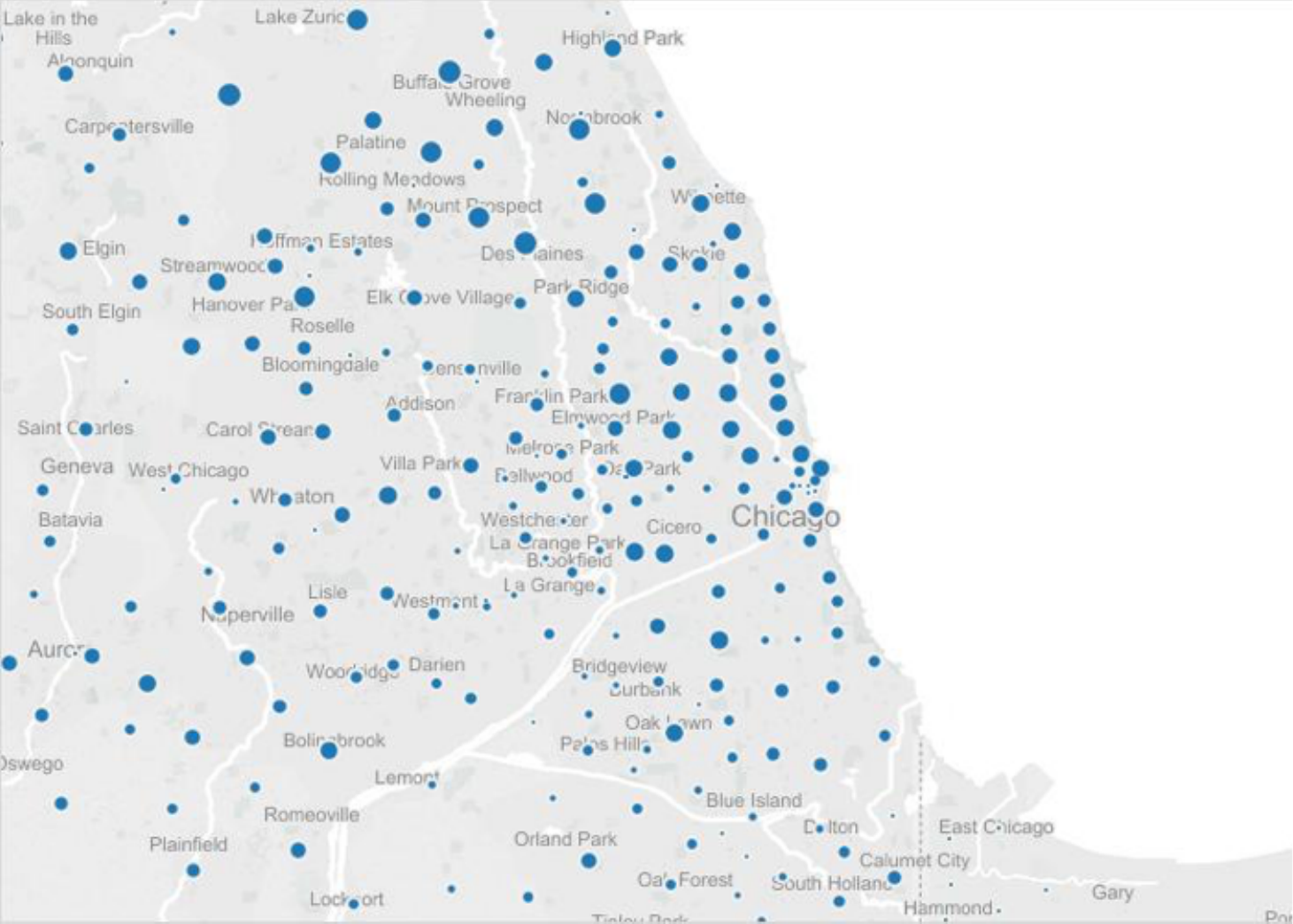}
	\caption{Geocoded Home Showings Data in the Chicagoland Area. Circle is proportional to the number of home showings. }
	\label{fig:map}
\end{figure}

The map above represents the volume of home showings in the ShowingTime dataset for Chicago-area home showings. Circles denote zip codes in which the homes are located and are scaled to increase with an increasing concentration of home showings. 

We use auto-regressive time series model with  Fourier terms to model cycles on the market.  When we forecast the weekly data, Fourier terms are a useful technique for dealing with cyclic data. In the case of weekly observations, the seasonal period is long and non-integer (there are $365.25/7 = 52.18$ weeks in a year), so a typical periodic forecasting ARIMA model or exponential smoothing do not appropriately treat such cyclicity even when 52 is used as an approximation of number of observations per year. Further, there can be cycles with different length, such as quarterly or monthly cycles. Fourier terms in the linear forecasting model allow for dealing with non-integer cycle lengths and multiple cycles of different lengths.  The general formula for the model is as follows 
\[
y_t = bt + \phi y_{y-1} + \beta^Tx + \sum_{j=1}^{K}\left(a_j \sin\left(\dfrac{2\pi j t}{52.18}\right) + b_j \cos\left(\dfrac{2\pi j t}{52.18}\right) \right) + \eta_t.
\]

Both Fourier parameters $\beta$'s and auto-regressive parameter can be estimated using quantum algorithms. Specifically, for estimating $\beta$'s, we use Quantum Fourier Transform (QFT), which is the  quantum implementation of the discrete Fourier transform over the amplitudes of a wavefunction. 

\subsection{Quantum FFT}
The quantum Fourier transform is naturally calculated in a quantum measurement system as it is already expressed in rotational form. The QFT takes a ``position'' state $|x\rangle$ to the corresponding momentum state $|p\rangle$ and is defined as follows
\[
\mathrm{QFT}_q(x) = \frac{1}{\sqrt{q}}\sum_{p=0}^{q-1} \exp(2\pi i p x/q)|p\rangle,
\]
where $q$ is the dimension of the systems Hilbert space. See \cite{weinstein2001implementation} for further discussion. In terms of gates, the two but QFT corresponds to the unitary operator
\[
\mathrm{QFT}_4 = 
\frac{1}{2}\begin{pmatrix}
1 & -1 & 1 & 1 \\
1 & i & -1 & -i \\
1 &-1 & 1 & -1 \\
1 & -i & -1 & i
\end{pmatrix}
\] 

The quantum Fourier transform acts on a quantum state $\vert X\rangle = \sum_{j=0}^{N-1} x_j \vert j \rangle$ and maps it to the quantum state $\vert Y\rangle = \sum_{k=0}^{N-1} y_k \vert k \rangle$ according to the formula
\[
	y_k = \frac{1}{\sqrt{N}}\sum_{j=0}^{N-1}x_j\omega_N^{jk}.
\]

We estimate the auto-regressive parameter $\phi$ using moment-based approach that relies on  Yule-Walker equations. This approach requires solving a system of linear equations that we solve using Quantum Linear Solver. Unlike well-known  HHL algorithm \cite{harrow2009quantum}, VQE does not require sparse matrix, which is the case in the problem of parameter estimation. 

The inputs into this algorithm are the matrix  $A$, which we have to decompose into a linear combination of unitaries with complex coefficients $A \ = \ \displaystyle\sum_{n} c_n \ A_n$.
 Then VQE, as any other variational quantum algorithm \cite{cerezo2021variational}, constructs a quantum cost function, which can be evaluated with a low-depth parameterized quantum circuit, then output to the classical optimizer. This allows us to search a parameter space for some set of parameters $\alpha$, such that $|\psi(\alpha)\rangle \ 
 = \|\textbf{x}\rangle / || \textbf{x} || $, where $|\psi(k)\rangle$  is the output of out quantum circuit corresponding to some parameter set $k$.

\begin{figure}[H]
	\centering
	\includegraphics[width=0.6\linewidth]{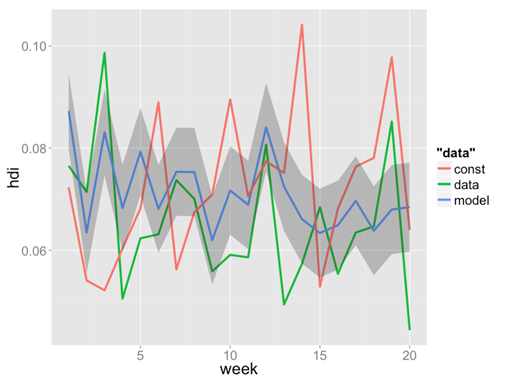}
	\caption{Time Series Plots of Showings by Date. 20-week forecast using the time series with exogenous predictors and Fourier terms. }
	\label{fig:fourier}
\end{figure}

Out-of-sample forecast for our model is shown in Figure~\ref{fig:fourier}. 

\section{Discussion}
Whilst quantum hardware in its infancy, the theoretical and quantum algorithms are well-developed. Our goal is to show how to apply these to machine learning and AI. 
In many cases, quantum von Neumann measurement provide an exponential speedup in computations. We outline a unified framework for quantum machine learning algorithms. We embed regression and classification methods, Bayesian learning, MCMC simulations \cite{hammersley1964monte} and deep learning into quantum measurement systems. As with ML algorithms, quantum ML uses kernel methods and feature extraction. Again storage and calculations of inner products and matrix inversions  provide exponential increases in speed.

Harnessing the power of quantum computing is one of the challenges in the $21$st century.


\bibliographystyle{apalike}
\bibliography{ref}

\section{Appendix A: Quantum Algorithms}

There are many quantum  algorithms which include Grover's algorithm, quantum Bayesian Shor's factorisation algorithm which has implications in cryptography and quantum versions of standard machine learning algorithms.

\paragraph{Quantum Cryptography: Shor's Algorithm}
Shor's quantum factoring algorithm has implications in cryptography and the strong Church-Turing thesis, see \cite{wang2022quantum} for further discussion. For classical computer, finding all price factors of a given number is known to be $O(\exp(n^{1/3}\log^{2/3}n))$. \cite{shor1994algorithms} showed theoretically that quantum computer can factorize any number in $O(n^2\log n\log\log n)$ operations.  This has implications for Quantum Encryption.

\cite{harrow2020Small} shows how to use classical computers for data pre-processing and quantum computers for sampling and optimisation required to build a predictive model. This paper addresses the problem of loading a dataset onto a quantum computer. Computationally, this task is very expensive, however a pre-processing technique can be used to reduce the size of the training data set required by a quantum computer. Practically, also shows how the Grover algorithm can be used to solve several computational problems that arise in machine learning using a reduced data set. The problem is formulated as the following minimization problem
\[
\arg\min_{y \in Y} \sum_{x\in X} f(x,y)
\]
Here $Y$ indexes the set of candidate models, $X$ is the training data set and $f$ is a loss function. 

\vspace{0.15in}

\noindent{\bf Grover's Algorithm}
As input for Grover's algorithm, suppose we have a function 
\[f: \{0,1,\ldots,N-1\} \to \{0,1\}.\] In the ``unstructured database'' analogy, the domain represent indices to a database, and $f(x) = 1$ if and only if the data that $x$ points to satisfies the search criterion. We additionally assume that only one index satisfies $f(x^\star) = 1$, and we call this index $x^\star .$ Our goal is to identify $x^\star$.

\vspace{0.15in}

\noindent{\bf Harrow's Algorithm.} This applies Grover's algorithm to the fundamental estimator of Bayesian Machine learning, the MAP estimator, via the optimisation problem, 
\[
{\rm arg min}_{ y \in Y } \;  \sum_{x\in X} f(x,y) + \lambda \phi(y)
\]
where $ \phi(y) $  is a regularisation penalty and $ f( \cdot , \cdot ) $ is an empirical loss function. See \cite{peng2020Simulating} for further discussion.

\vspace{0.15in}
\noindent{\bf Quantum Monte Carlo: Annealing and Tunneling}
 This allows a  speed up of traditional Monte Carlo algorithms based on the notion of quantum tunneling. 
As with all quantum measurement there is a preparation of states, denoted by $| \psi \rangle$, then a quantum device is used for evolution of states allowing the 
application of unitary transformation $U$. 
See Wang (2011, 2016, 2022) for further discussion and algorithmic details.
Quantum annealing utilizes the physical process of a quantum system whose lowest energy, or equivalently, a ground state of the dynamic system, gives a solution to the posed
Monte Carlo problem via the solution of the Schr\"odinger equation.

There are many applications of quantum algorithms. One area of application is in quantum sensing.
\paragraph{Quantum Sensing} \cite{degen2017} Quantum Sensing is an advanced sensor technology that  improves the accuracy of how we measure, navigate, study, explore, see, and interact with the world around us by sensing changes in motion, an
\end{document}